# Image quality enhancement of embedded holograms in holographic information hiding using deep neural networks

Tomoyoshi Shimobaba [1], Sota Oshima [1], Takashi Kakue [1], and Tomoyoshi Ito [1]

[1]*Graduate School of Engineering, Chiba University, 1-33 Yayoi-cho, Inage-ku, Chiba 263-8522, Japan*

Holographic information hiding is a technique for embedding holograms or images into another hologram, used for copyright protection and steganography of holograms. Using deep neural networks, we offer a way to improve the visual quality of embedded holograms. The brightness of an embedded hologram is set to a fraction of that of the host hologram, resulting in a barely damaged reconstructed image of the host hologram. However, it is difficult to perceive because the embedded hologram's reconstructed image is darker than the reconstructed host image. In this study, we use deep neural networks to restore the darkened image.

## 1 Introduction

Steganography, which embeds other images in one image without the observer noticing, and watermarking, which allows the observer to know that a hidden image is embedded, has been developed for secure communication and copyright protection of images. These are collectively called information hiding. Recent advances in computer technology have made it possible to digitally generate holograms by simulating light interference and diffraction phenomena on a computer. Digitally generated holograms can be used for three-dimensional (3D) displays and projections. Information hiding technology for holograms has been developed.

Information hiding in holography can be classified into spatial domain and frequency-domain methods. In spatial domain methods, a hologram with a host image and a hologram with an embedded image are added together in the spatial domain [1]. An embedded image can be converted into an encrypted hologram using double random phase encoding [2]. To encrypt embedded holograms, fractional Fourier transforms [3–6], Fresnel transforms [7], and gyrator transforms [8] can be employed. A method of embedding 3D object-hiding holograms within host holograms recorded with 3D objects has been presented [9], as has a way of substituting low bits of a hosted picture with an embedded hologram [10–13]. A holographic micro information hiding [14] has been proposed to embed small unnoticeable hiding information into a host hologram using scale diffraction calculations [15–17]. A host hologram is transformed to the frequency domain by using mathematical transforms, and a hidden hologram is embedded in the frequency domain. The Fourier transform, discrete cosine transform, wavelet transform [18], and gyrator transform [19] can transform holograms into the frequency domain. In general, they are referred to as linear canonical transforms.

In this paper, we present a technique for embedding another hologram into a host hologram without compromising the reconstructed picture of the host hologram and for improving the image quality of the embedded hologram using deep neural networks (DNNs). In the suggested method, the brightness of an embedded hologram is set to a few percent of the brightness of a host hologram. The reconstructed image of the host hologram is barely degraded. However, the reconstructed image of the embedded hologram is darker than that of the host hologram, which makes it difficult to see. To restore the darkened image, we employ DNNs.

## 2   Proposed method

Figure 1 shows a setup of the holographic information hiding used in this study. A host image and embedding image are placed opposite to each other on the hologram.

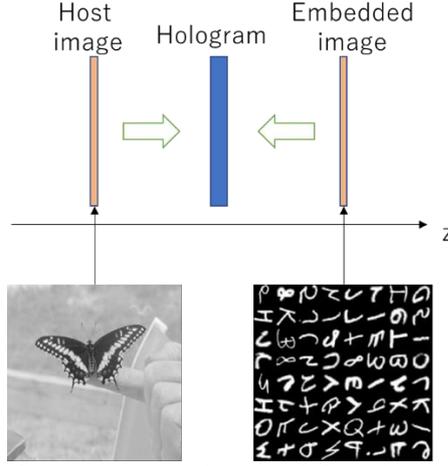

Fig. 1 Setup for the proposed holographic information hiding.

By arranging them, the other reconstructed images are blurred in the reconstructed image to be gazed at, making it easier to see the reconstructed image. In this study, we used two-dimensional images for a host and embedded images, but 3D images are also acceptable. The proposed hologram can be calculated as

$$u(x) = \wp_{z_1}(u_h(x)) + \alpha \wp_{z_2}(u_e(x)), \quad (1)$$

where $u(x)$ denotes the final hologram, $\wp_z$ denotes a diffraction calculation at a propagation distance of $z$, $u_h(x)$ and $u_e(x)$ represent a host image and embedded image, respectively, and $\alpha$ denotes the strength of the embedding. Any diffraction calculation can be used, but here we used the angular spectral method [20]. Host holograms and embedded holograms are normalized.

If7 this final hologram is reconstructed at a distance of $z_1$, the host image can be reconstructed, and if it is reconstructed at a distance of $z_2$, the embedded image can also be reconstructed. Figure 2 shows an example of reconstructed images. For the reconstruction calculation, the angular spectrum method is applied. In theory, both reconstructed images contain the other reconstructed images; however, by setting to a small value of $\alpha$, the image quality of the host images can be preserved. We used $\alpha$=0.04 in this case. Figure 2(a) depicts a reconstructed host picture at $z_1$ distance and Fig. 2(b) depict an enlarged view of the area containing an embedded image. Figure 2(e) depicts an even more magnified perspective. As can be shown, the reconstructed embedded picture has essentially no effect on the reconstructed host image. Figure 2(c) shows the reconstructed embedded image at a distance of $z_2$. Figure 2(d) shows an enlarged view of the area containing the embedded reconstructed image. Figures 2(f) and 2(g) show enlarged views of the areas where "8" (rotated 90 degrees) and "0" are recorded in the reconstructed embedded image, respectively. By zooming in, the images can be seen dimly, but it is difficult to observe.

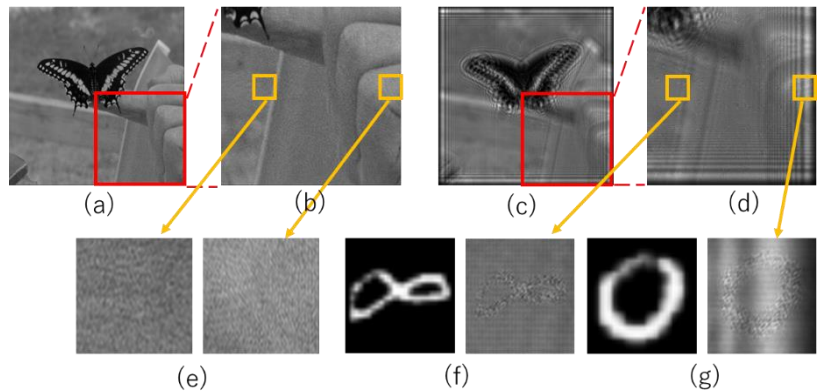

Fig. 2 Reconstructed host and embedded images: (a) and (b) reconstructed host image and its magnified view, (c) and (d) show the reconstructed embedded image and its magnified view, (e) magnified view of the reconstructed embedded image in the reconstructed host image, and (f) and (g) magnified view of the reconstructed embedded image and its original embedded image.

In this study, we use DNNs to restore this hard-to-see reconstructed embedded image. Figure 3 shows the proposed method. If an entire reconstructed embedded image to the DNN at once, it requires many network parameters. Instead, we divide the reconstructed embedded image into blocks, and the DNN performs image restoration for the blocks. This way, the training time of the DNN can be shortened, and the image restoration can be done independently of the hologram resolution.

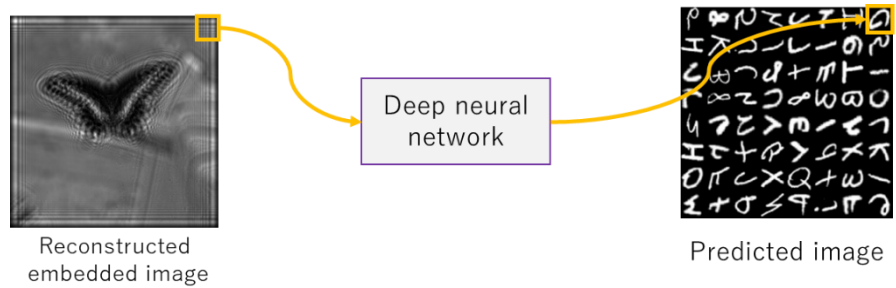

Fig. 3 Proposed method: DNN restores the reconstructed embedded image block by block instead of restoring the entire image at once.

In this study, we used U-Net [21] and ResNet [22] as the network structure of DNN. We compare the results of each network in the next section. Figure 4 shows the structure of U-Net. U-Net is a DNN that uses skip connection, which is a method of passing high-resolution information before down-sampling directly to the up-sampling layer, thus preserving information that is lost in down-sampling. In Fig. 4, "Conv" is the convolution layer, "MaxPooling" is the maxpooling layer, and "UpSampling" is the upsampling layer, which up-samples the input by a factor of two. The notation (H × W × C) of each layer indicates the height (H), width (W), and a number of channels (C) of the image. The structure of the "Block" is depicted in Fig. 4(b). The "Block" is made up of batch normalization and convolutional layers. In the convolutional layers, the notation "Conv(F, (S × S))" denotes that F is the number of filters and S × S is the filter size. The stride width was 1 × 1, and padding was used to make the output image the same size as the input image. The rectified linear unit (ReLU) function was used as the activation function. The convolutional layer just before the output layer in Fig. 4(a) was set to have a filter size of 1 × 1 and a number of filters of 1.

Figure 5 shows the structure of ResNet. ResNet is a DNN that introduces a shortcut connection between the layers. This shortcut connection can avoid the gradient vanishing problem during training even if the DNN has many layers. The module "A" in Fig. 5 consists of a batch normalization layer, a "Block," and an additional layer. The structure of "Block" consists of the convolutional layers and the batch normalization layers. In all the convolution layers, the stride was 1 × 1, and the activation function was the ReLU function.

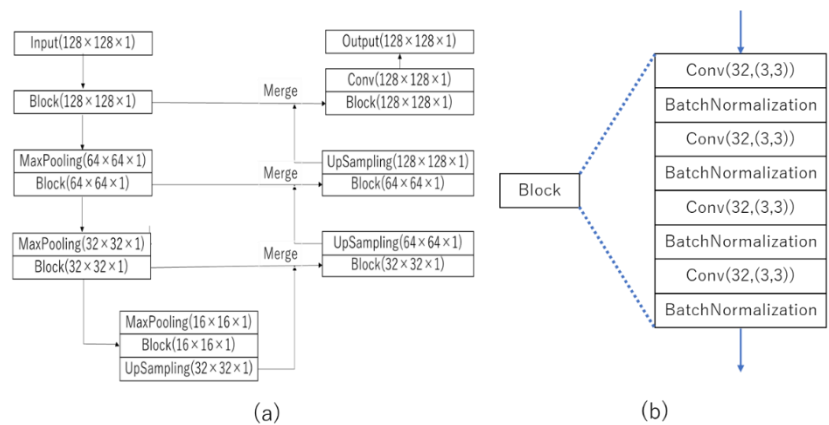

Fig. 4 U-Net.

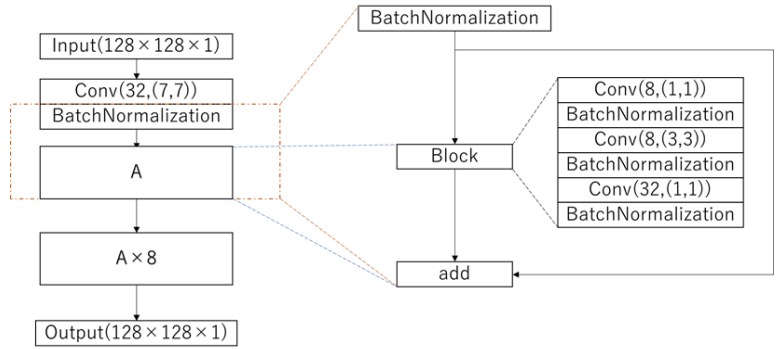

Fig. 5 ResNet.

## 3 Results

      This study needs to prepare a dataset containing holograms recorded host and embedded images and original embedded images for training the DNNs. Figures 6(a) and 6(b) show examples of host and embedded images. Host images were randomly extracted from images contained in caltech256. We resized and grayscaled these images to 2048 × 2048 pixels. The embedded image can be a simple image for applications that embed copyright protection or simple hidden messages. For the embedded images, we used the MNIST dataset of handwritten numeric images. MNIST images having 28 × 28 pixels were resized to 128 × 128 pixels, and the resized MNIST images were arranged in 8 × 8 horizontally and vertically to generate the embedded image. We created a text-free empty region in the embedded image. This is to prevent the DNNs from mistakenly restoring the area that does not contain the embedded information.

      Figures 6(c) and 6(d) show reconstructed host and embedded images from holograms calculated by Eq. (1). The recording distance was set to −0.4 m for the host image and 0.4 m for the embedded image. We used a wavelength of 633 nm and a sampling interval of 10 μm for the hologram, and we applied random phases to distribute the information of the embedded image over the entire hologram. We cannot see the embedded image in the reconstructed host and embedded images. We included 300 of these images in our data set.

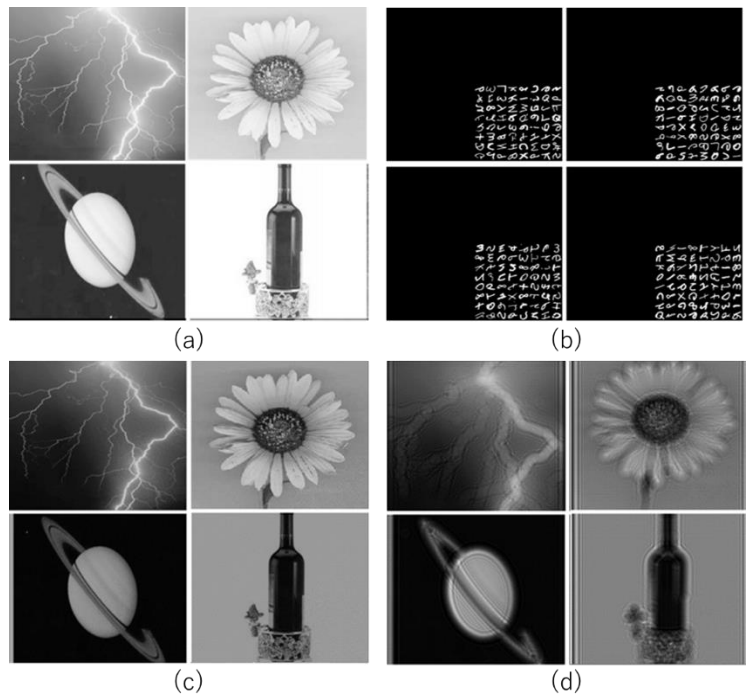

Fig. 6 Original host and embedded images and their reconstruction from the holograms: (a) original host images, (b) original embedded images, (c) host reconstructed images, and (d) embedded reconstructed images.

The DNNs in Fig. 4 and Fig. 5 require input images of 128 × 128 pixels. Therefore, we divided the 2048 × 2048-pixel reconstructed embedded image into 128 × 128-pixels images. We can utilize a set of 76,800 images with 128 × 128 pixels because the dataset has 300 image sets. We used 75,520 and 1280 image sets for training and validation, respectively. The imagess are input to the DNNs. We trained the DNNs by the loss function of the mean squared error between its output and the original embedded image, which is the ground-truth image. We used ADAM for the optimizer and the batch size of 50.

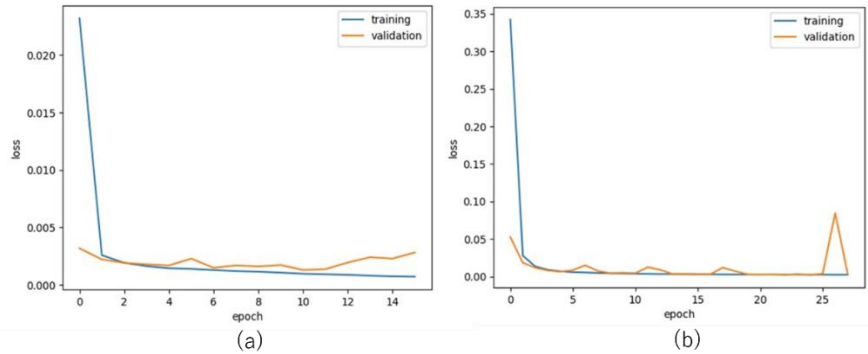

Fig. 7 Training and validation curves of U-Net (Fig. 4) and ResNet (Fig. 5).

Figure 7 shows U-Net's training and validation curves (Fig. 4) and ResNet (Fig. 5). We used early stopping. As a result, the number of epochs differed since the training automatically terminated when the loss values stopped changing. Since the loss values of U-Net were one order of magnitude lower than that of U-Net, we show only the results of U-Net hereafter.

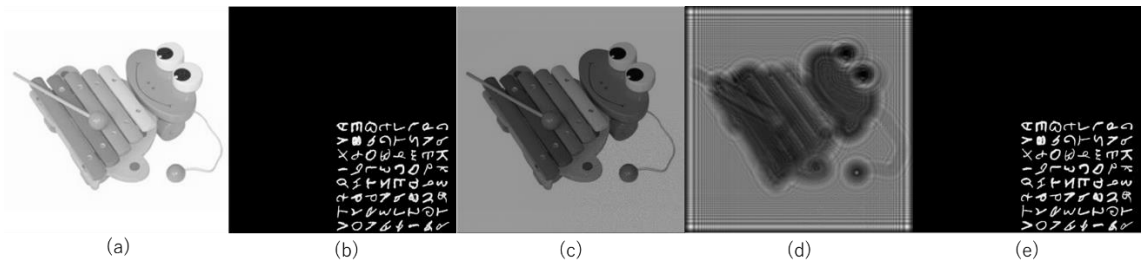

Fig. 8 Original host image, embedded image, and their reconstructed images: (a) and (b) are original host image and embedded image, (c) and (d) are the reconstructed host image and reconstructed embedded image, and (e) is the embedded image restored by the DNN.

Figure 8 shows the results when using the trained U-Net. Figures 8(a) and 8(b) show the original host image and embedded image. Figures 8(c) and 8(d) show the reconstructed host image and reconstructed embedded image. By inputting Fig. 8(d) into the DNN, we obtained Fig. 8(e). Figure 8(e) is comparable to the original embedded image.

Finally, in Fig. 9, we show the results of using embedded images, which are different from the MNIST images. The DNN employed here is the same as in Fig. 8. Figures 9(a) and 9(b) depict the original host and embedded pictures, respectively, whereas Figs. 9(c) and 9(d) depict the reconstructed host and embedded images restored by the DNN. The peak signal-to-noise ratio (PSNR) and structural similarity were used to compare the image quality of the original embedded image and the restored embedded image (SSIM). The PSNR and SSIM were 29.6 dB and 0.98, respectively, indicating that the restored image was of sufficiently high quality.

The reference [23] found the optimal the weighting factor analytycally. Compared to their weight factor, we were able to reduce our weight factor by one order of magnitude. This means that the effect on the reconstructed host image can be reduced. The reference you suggested used linear signal processing to reconstruct the host image and hidden image, but we used nonlinear signal processing (deep learning) to recover the hidden image, so we were able to use fewer weight coefficients.

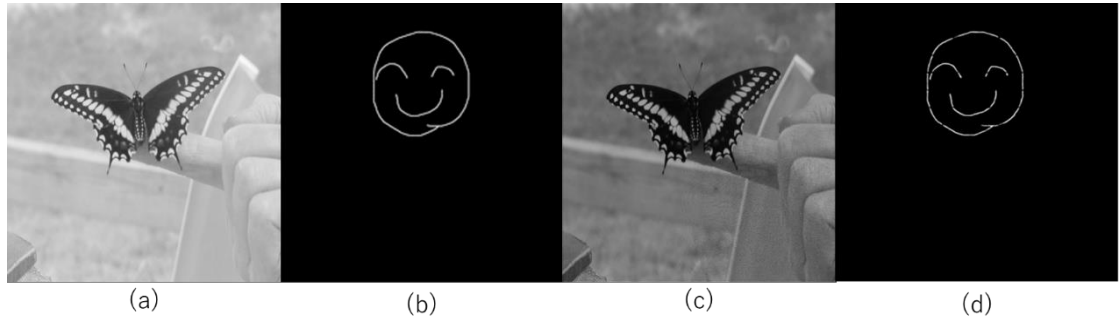

Fig. 9 Original host image, embedded image, and their reconstructed images: (a) and (b) are original host image and embedded image, (c) is the host reconstructed image, and (d) is the embedded image restored by the DNN.

## 4    Conclusions

In the proposed holographic information hiding method, the brightness of embedded holograms was set to 4% of that of the host holograms. These two holograms were superimposed in the spatial domain to produce the final holograms. The visual quality of the reconstructed host images from these holograms was excellent. The recreated embedded images, on the other hand, became black and difficult to view. To improve the darkened images, we employed DNNs to improve the image quality. We could obtain a restored embedded image with the PSNR and SSIM of 29.6 dB and 0.98, respectively,

**Acknowledgments**

This research was funded by Japan Society for the Promotion of Science (19H04132, 19H01097).

**References**

[1] Aoki Y, Watermarking technique using computer-generated holograms, Electron. Commun. Jpn. 84 (2001) 21–31.
[2] Kishk S, Javidi B, Information hiding technique with double phase encoding, Appl. Opt. 41 (2002) 5462–5470.
[3] Unnikrishnan G, Joseph J, and Singh K, Optical encryption by double-random phase encoding in the fractional Fourier domain, Opt. Lett. 25 (2000) 887-889.
[4] Unnikrishnan G, Singh K, Double random fractional Fourier domain encoding for optical security, Opt. Eng. 39 (2000) 2853-2859.
[5] Nishchal N V, Joseph J, Singh K, Fully phase encryption using fractional Fourier transform, Opt. Eng. 42 (2003) 1583-1588.
[6] Nishchal N K, Pitkäaho T, Naughton T J, Digital Fresnel hologram watermarking, in 9th Euro-American Workshop on Information Optics (WIO) (IEEE, 2010), 1–3.
[7] Situ G, Zhang J, Double random-phase encoding in the Fresnel domain, Opt. Lett. 29 (2004) 1584-1586.
[8] Yadav A K, Vashisth S, Singh H, Singh K, A phase-image watermarking scheme in gyrator domain using devil's vortex Fresnel lens as a phase mask, Opt. Commun. 344 (2015) 172-180.
[9] Kishk S, Javidi B, 3D object watermarking by a 3D hidden object, Opt. Express 11 (2003) 874–888.
[10] Hamam H, Digital holography-based steganography, Opt. Lett. 35 (2010) 4175–4177.
[11] Tsang P, Poon T C, Data-embedded-error-diffusion hologram, Chin. Opt. Lett. 12 (2014) 60017–60020.
[12] Tsang P, Poon T C, Chow Y T, Embedding image into a phase-only hologram, Opt. Commun. 341 (2015) 188–193.
[13] Hwang H E, Chang H T, Lie W N, Lensless optical data embedding system using concealogram and cascaded digital Fresnel hologram, J. Opt. Soc. Am. A 28 (2011) 1453–1461.
[14] Shimobaba T, Endo Y, Hirayama R, Hiyama D, Nagahama Y, Hasegawa S, Sano M, Takahashi T, Kakue T, Oikawa M, Ito T, Holographic microinformation hiding, Appl. Opt. 56 (2017) 833-837.
[15] Muffoletto R P, Tyler J M, Tohline J E, Shifted Fresnel diffraction for computational holography, Opt. Express 15 (2007) 5631-5640.
[16] Odate S, Koike C, Toba H, Koike T, Sugaya A, Sugisaki K, Otaki K, Uchikawa K, Angular spectrum calculations for arbitrary focal length with a scaled convolution, Opt. Express 19 (2011) 14268–14276.
[17] Shimobaba T, Kakue T, Okada N, Oikawa M, Yamaguchi Y, Ito T, Aliasing-reduced Fresnel diffraction with scale and shift operations, J. Opt. 15 (2013) 075405(5pp).
[18] Dehghan H, Safavi S E, Robust image watermarking in the wavelet domain for copyright protection, arXiv preprint arXiv:1001.0282 (2010).

[19] Vashisth S, Yadav A K, Singh H, Singh K, Watermarking in gyrator domain using an asymmetric cryptosystem, Proc. SPIE 9654 (2015) International Conference on Optics and Photonics 2015, 96542E.
[20] Matsushima K, Shimobaba T, Band-Limited Angular Spectrum Method for Numerical Simulation of Free-Space Propagation in Far and Near Fields, Opt. Express 17 (2009) 19662-19673.
[21] Ronneberger O, Fischer P, Brox T, U-net: Convolutional networks for biomedical image segmentation, International Conference on Medical image computing and computer-assisted intervention (2015).
[22] He K, Zhang X, Ren S, Sun J, Deep residual learning for image recognition, In Proceedings of the IEEE conference on computer vision and pattern recognition (2016) 770-778.
[23] Kim H, Lee Y H, Optimal watermarking of digital hologram of 3-D object," Opt. Express 13 (2005) 2881-2886.